\documentclass[letterpaper]{article} 
\usepackage{aaai24}  
\usepackage{times}  
\usepackage{helvet}  
\usepackage{courier}  
\usepackage[hyphens]{url}  
\usepackage{graphicx} 
\usepackage{natbib}  
\usepackage{caption} 
\usepackage{algorithm}
\usepackage{algorithmic}

\usepackage{newfloat}
\usepackage{listings}
\DeclareCaptionStyle{ruled}{labelfont=normalfont,labelsep=colon,strut=off} 
\floatstyle{ruled}
\newfloat{listing}{tb}{lst}{}
\floatname{listing}{Listing}
\usepackage{tabularray}
\usepackage[table]{xcolor}
\usepackage{colortbl}
\definecolor{lightgreen}{rgb}{0.56, 0.93, 0.56} 
\definecolor{lightred}{rgb}{1.0, 0.71, 0.76}
\newlength{\mybarlen}
\usepackage{multirow}
\usepackage{multicol}
\usepackage{calc}
\newcommand\Tstrut{\rule{0pt}{2ex}}         
\newcommand\Bst{\rule[-0.8ex]{0pt}{0pt}}   

\title{Beyond Training: Optimizing Reinforcement Learning Based Job Shop Scheduling Through Adaptive Action Sampling}
\author{
    Constantin Waubert de Puiseau,
    Christian Dörpelkus,
    Jannik Peters,
    Hasan Tercan, 
    Tobias Meisen
}
\affiliations{
    Institute for Technologies and Management of Digital Transformation\\

    Lise-Meitner-Strasse 27-31\\
    42119 Wuppertal\\
    \{waubert, christian.doerpelkus, jpeters, tercan, meisen\}@uni-wuppertal.de
}

\usepackage{bibentry}

\begin{document}

\maketitle

\begin{abstract}
Learned construction heuristics for scheduling problems have become increasingly competitive with established solvers and heuristics in recent years. In particular, significant improvements have been observed in solution approaches using deep reinforcement learning (DRL). While much attention has been paid to the design of network architectures and training algorithms to achieve state-of-the-art results, little research has investigated the optimal use of trained DRL agents during inference. Our work is based on the hypothesis that, similar to search algorithms, the utilization of trained DRL agents should be dependent on the acceptable computational budget. We propose a simple yet effective parameterization, called $\delta$-sampling that manipulates the trained action vector to bias agent behavior towards exploration or exploitation during solution construction. By following this approach, we can achieve a more comprehensive coverage of the search space while still generating an acceptable number of solutions. In addition, we propose an algorithm for obtaining the optimal parameterization for such a given number of solutions and any given trained agent. Experiments extending existing training protocols for job shop scheduling problems with our inference method validate our hypothesis and result in the expected improvements of the generated solutions.
\end{abstract}

\section{Introduction}
Research on self-learning algorithms for scheduling problems has significantly increased in recent years, driven primarily by advancements in algorithms and the availability of more affordable and powerful computing units in both industry and academia. In academia, the standardized job shop scheduling problem (JSSP) \cite{Pinedo.2016} is the dominant object of study for its straightforward logical constraints and comparability thanks to public benchmarks and libraries \cite{vanHoorn.2018, WaubertdePuiseau.2023b}. While in some successful DRL-based solution methods agents steer existing search algorithms \cite{Zhang.20.11.2022, Ni.2021} and solvers \cite{Tassel.2023}, in a majority, agents autonomously generate solution schedules imitating a construction heuristic \cite{zhang.2020, park.2021, vanekeris.2021, Tassel.2022b}.

In these DRL-based construction heuristics, agents make iterative decisions to schedule the next unscheduled operation until the entire schedule is complete. For each decision, an action is sampled from the agent's action vector that consists of logits representing the current operation prioritization. Depending on how actions are sampled from the vector, different solutions for a JSSP are obtained.
For our purposes, we differentiate between the following sampling strategies:
\begin{itemize}
    \item \textbf{Deterministic sampling:} by greedily applying \textit{argmax} policy logits we obtain one deterministic solution per problem instance.
    \item \textbf{Stochastic sampling:} sampling from the given distribution, where the likelihood of each action is given by the corresponding logit. 
    \item \textbf{Exploitative/Explorative sampling:} during exploitative sampling, the likelihoods of likely actions increase even more; during explorative sampling, the likelihoods of all actions are assimilated. We achieve both through the $\delta$-sampling which we present in this paper. 
\end{itemize}

\begin{figure}[ht]
  \centering
  \includegraphics[width=0.8\linewidth]{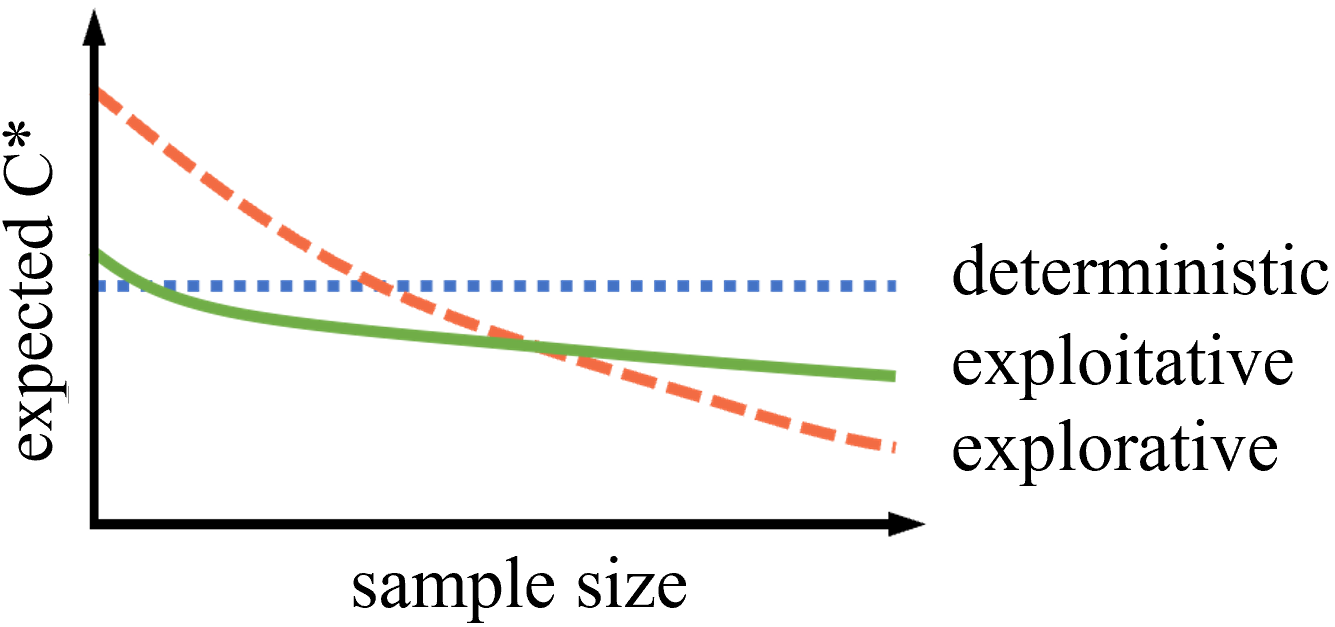}
  \caption{Expected minimal makespans $C^{*}$ over sampling size for different sampling strategies}
  \label{fig:hypothesis}
\end{figure}

In their extremes, exploitative sampling converges towards deterministic sampling whereas explorative sampling converges towards sampling from a uniform distribution. Stochastic sampling falls in between. 
In the context of scheduling, it has only very recently been shown that stochastic sampling can produce better-than-greedy schedules even with a relatively small number of sampled solutions \cite{Iklassov.09.06.2022b}. Motivated by this finding, our goal is to identify the optimal sampling method for any given trained agent and the desirable number of samples to be evaluated. During inference, the objective is to sample in such a way that among all sampled solutions, the completion times $C\textsubscript{max}$, also known as makespans, the shortest is as short as possible. To avoid confusion with the common subscript $max$, and for brevity, we refer to this minimum value as $C^{*}$ throughout this paper.
According to the proposed nomenclature, we formulate our working hypothesis as follows:
\begin{enumerate}
    \item The expected $C^{*}$ is a function of the sampling strategy and number of sampled solutions, i.e., the sample size.
    \item For a fixed (non-deterministic) sampling strategy, $C^{*}$ remains equal or decreases with increasing sample size.
    \item For a small sample size, strongly exploitative sampling strategies statistically lead to lower $C^{*}$ compared to very explorative strategies. In other words, it is unlikely to find a local and better-than-greedy minimum when deviating more strongly from the greedy solution.
    \item Conversely, for a large sample size, strongly explorative sampling strategies statistically lead to better $C^{*}$ compared to exploitative strategies, since a wider solution space is covered.
\end{enumerate}
Figure~\ref{fig:hypothesis} summarizes our hypothesis. Moreover, if the hypothesis is valid, there should exist a sampling strategy that minimizes the expected $C^{*}$ for a given agent and sample size.

Building upon our hypothesis, we introduce a new sampling method and validate its effectiveness on agents trained with two different state-of-the-art training protocols, testing them on JSSP instances of different sizes and with different sample sizes. Our study leads to the following valuable contributions:

\begin{enumerate}
    \item A new simple but effective tree-search-like solution generation algorithm for JSSP based on action sampling
    \item An algorithm for the identification of the most effective parameterizations of action sampling methods given a trained agent and a number of samples
    \item A validation of our algorithm on varios JSSP instances of different sizes as well as on  benchmark instances of \citet{Taillard.1993}. 
\end{enumerate}

The remainder of this paper is structured as follows: After discussing related work, we describe our methods and experiments. We then present and discuss the experimental results, before drawing conclusions and giving an outlook on future work.

\section{Related Work}\label{related_work}

\textbf{DRL-based construction heuristics.}
Early successful attempts to solve combinatorial optimization problems with DRL in 2016 \cite{bello.2016} were followed by a growing wave of research tackling the JSSP \cite{Panzer.2022}. Recent research focuses on effective problem size agnostic neural network architectures, including graph embeddings \cite{zhang.2020, park.2021} and recurrent networks \cite{Iklassov.09.06.2022b}. Another trend is the use of curriculum learning \cite{Iklassov.09.06.2022b, WaubertdePuiseau.2023}.
Interestingly, only \citet{Iklassov.09.06.2022b} sample multiple solutions stochastically, although the possibility to obtain better solutions through sampling in combinatorial problems was already established by \citet{WouterKool.2018}. However, to the best of our knowledge, our study is the first to focus on improving sampling methods for generating JSSP solutions using trained agents.

\textbf{Policy-guided tree search.}
Stochastic sampling from a learned policy is a common concept to guide tree search algorithms \cite{Orseau.}. The most common search algorithm for this integration is Monte-Carlo tree search (MCTS). Famously, this combination was leveraged in AlphaGo \cite{silver.2016} and its successors. Therein, the expansion of nodes is determined by a linear combination of learned Q-values or action predictions and the visit count of states. Lately, the approach has been successfully transferred to scheduling as well \cite{oren.2021, Kumar.2021}. However, MCTS comes at a high computational expense that is accepted in many cases because it provides good expected values even in stochastic environments that are infeasible to approximate analytically. For the deterministic JSSP, however, the computational budget may be used more efficiently by sampling multiple whole trajectories depth-wise from start to finish, as done in this study.

\textbf{Solution space coverage.}
The question of how closely to adhere to a learned policy is captured in the exploration-exploitation tradeoff. However, this tradeoff is usually exclusively considered during training. Therefore, methods like epsilon-greedy sampling \cite{Sutton.2018} or added loss terms on the action confidence through entropy or curiosity terms \cite{schulman.2017, DeepakPathak.2017} do not transfer to inference. 
A noteworthy technique to fine-tune exploration during inference in MCTS is to use an exponent across the action vector \cite{Wang.2021h, Shen.13.02.2018}. The primary rationale is a de- or increased effect of action priors compared to visit counts, but a secondary effect is a (dis-)alignment of values in the action vector itself. Our method is inspired by this idea and leverages the secondary effect.

For combinatorial optimization problems, depth-wise search with DRL is mostly done by stochastic policy sampling \cite{Iklassov.09.06.2022b, WouterKool.2018, Orseau.}. One exception is beam-search, which ensures a fixed tree width $k$ during search \cite{Vinyals.692015}. Known extensions make use of problem symmetries \cite{.2020d} and entropy in the action vector for the solution space expansion \cite{Tassel.2022b}. \citet{Iklassov.09.06.2022b} compared stochastic sampling and beam-search variations, finding that stochastic sampling outperforms the other methods. However, they only reported results on a single sample size. We expand this study with a new sampling method which fine-tunes sampling for each combination of trained agent and desired sample size.

\section{Method and Experiments}

\subsection{$\delta$-Sampling}
Our proposed sampling strategy offers the possibility to adjust the balance between exploitative and explorative sampling. To this end, we manipulate the action vector $a$ predicted by a trained agent to obtain $a_{sample}$ by exponentiating the predicted logits with the discrimination exponent $\delta$ and then re-scale the vector as in equation \ref{eq:deltasampling}. 

\begin{eqnarray}\label{eq:deltasampling}
a_{sample} = \frac{1}{{\sum_{i} a_i^\delta}} \cdot a^\delta
\end{eqnarray}

The effect is that for $\delta > 1$, relative differences between logits are emphasized favoring exploitation, whereas for $\delta < 1$ differences are decreased favoring exploration. For $\delta = 1$, logits remain unchanged, defaulting to stochastic sampling. We call this procedure $\delta$-sampling. It easily scales to any problem size and allows to precisely adjust the sampling behavior.

\begin{figure}[ht]
  \centering
  \includegraphics[width=1\linewidth]{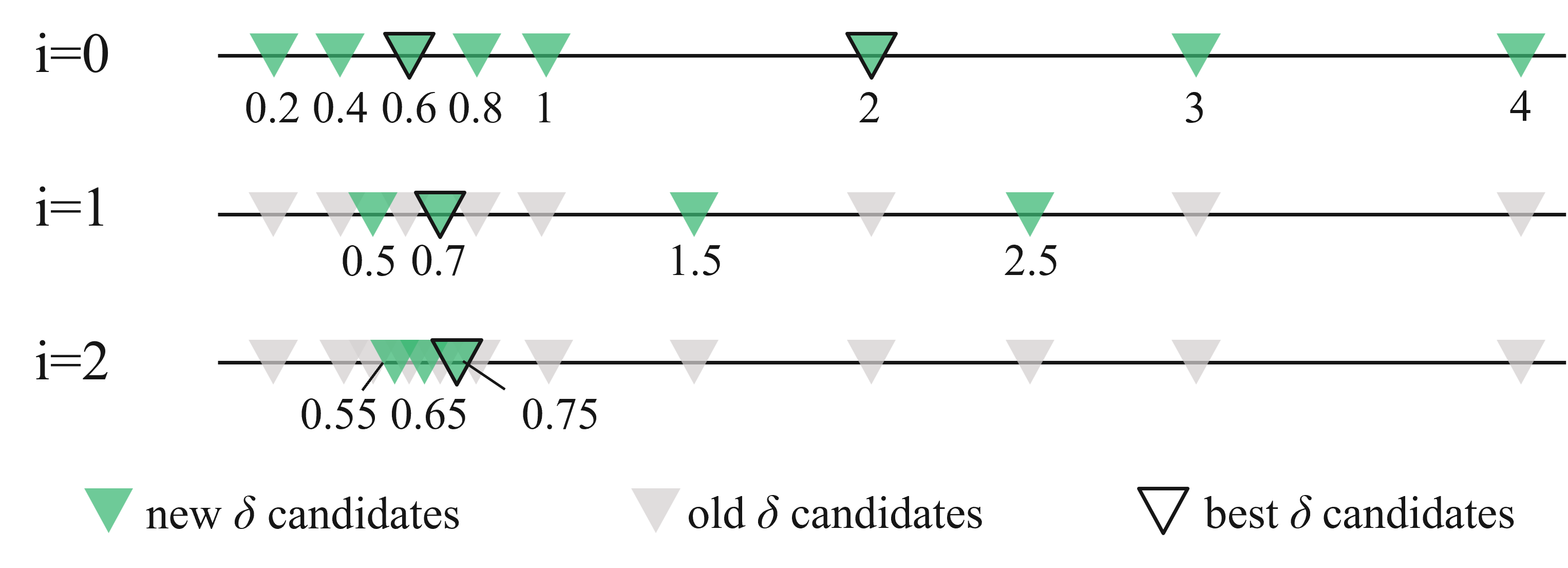}
  \caption{Three iterations of searching the optimal $\delta$ value}
  \label{fig:delta_algorithm}
\end{figure}

Given a base model and sample size, we are looking for  $\delta^{*}$ that minimizes $C^{*}$. We expect the functions of $C^{*}(\delta^{*})$ to be smooth and have a single minimum based on our hypothesis. Hence, our algorithm starts with a grid search that iteratively decreases the distance between the two most promising candidate values for $\delta$ to neighboring candidates by 50\%. An example is illustrated in Figure \ref{fig:delta_algorithm}. In the first iteration (i=0), we sample with each candidate $\delta$ from a list of candidates and the desired sample size on validation JSSP instances i\textsubscript{$\delta$}. The initial list of candidates is a hyperparameter that has to be set. Sampling returns the average of $C^{*}$ across all i\textsubscript{$\delta$} per candidate. Then, additional candidates are added between the two most promising and their adjacent candidates, rounding to two decimal places. If the most promising candidates lie at minimum or maximum $\delta$ values, the initial interval is extended by half of the largest distance between adjacent candidates. The function repeats until a specified number of iterations or minimal distance between adjacent candidates is reached. 

\subsection{Experimental Setup}

\textbf{JSSP formalization.} This work deals with the common JSSP formalization in which $J$ jobs, consisting of $O$ operations are processed on $M$ machines. Each job visits each machine exactly once in a predefined order, such that $O=M$. Additionally, each operation has a specified processing time and may not be interrupted once started (no preemption). Machines can only process one job at once (no overlap) and setup-times are ignored. The notation of problem sizes follows the structure $J$x$M$, e.g., a 20x15 instance consists of 20 jobs á 15 operations on 15 machines. 

\textbf{Base Models.} We perform experiments on two base model designs that are adapted from \citet{zhang.2020} and \citet{Iklassov.09.06.2022b}, two competitive and often-referenced representatives of DRL-based construction heuristics. We refer to the base models as \textit{L2D} \cite{zhang.2020} and \textit{L2G} \cite{Iklassov.09.06.2022b} in reference to their publication titles. From \textit{L2D}, we can extract the openly accessible pre-trained graph neural network weights for each respective problem size 6x6, 15x15 and 20x20, except for 100x20, where we resort to the 30x20 model as done in the original publication. 
The second base model, \textit{L2G}, features a neural network architecture that uses a set2set recurrent layer \cite{Vinyals.19.11.2015} to be problem size agnostic. Since no trained models are publicly available, we retrain our base models according to the reported protocol, one for each problem size, with the following curricula: for the 15x15 and 6x6 JSSP models, we choose adaptive curriculum learning and train on 6x6 and 15x15 JSSP instances. For the 20x20 and 100x20 base model, we use the RASCL method \cite{Iklassov.09.06.2022b} and train on 6x6, 15x15, and 20x20 JSSP instances.

\textbf{JSSP Instances.} 
We test our sampling method on four problem sizes. 15x15, 20x20, and 100x20 are common benchmark sizes in the operations research domain. 6x6 is also common in the literature since it is useful for the visual interpretations of results in Gantt charts. To avoid overfitting on test data, we create separate JSSP instances for training and for the search of suitable sampling parameterizations. These Taillard-like instances are created with separate random seeds, assigning random perturbations of machine sequences to each job and sampling processing times uniformly from the interval [1, 99].

\section{Results}
\subsection{Hypothesis Validation}
\begin{figure}[ht]
  \centering
  \includegraphics[width=0.9\linewidth]{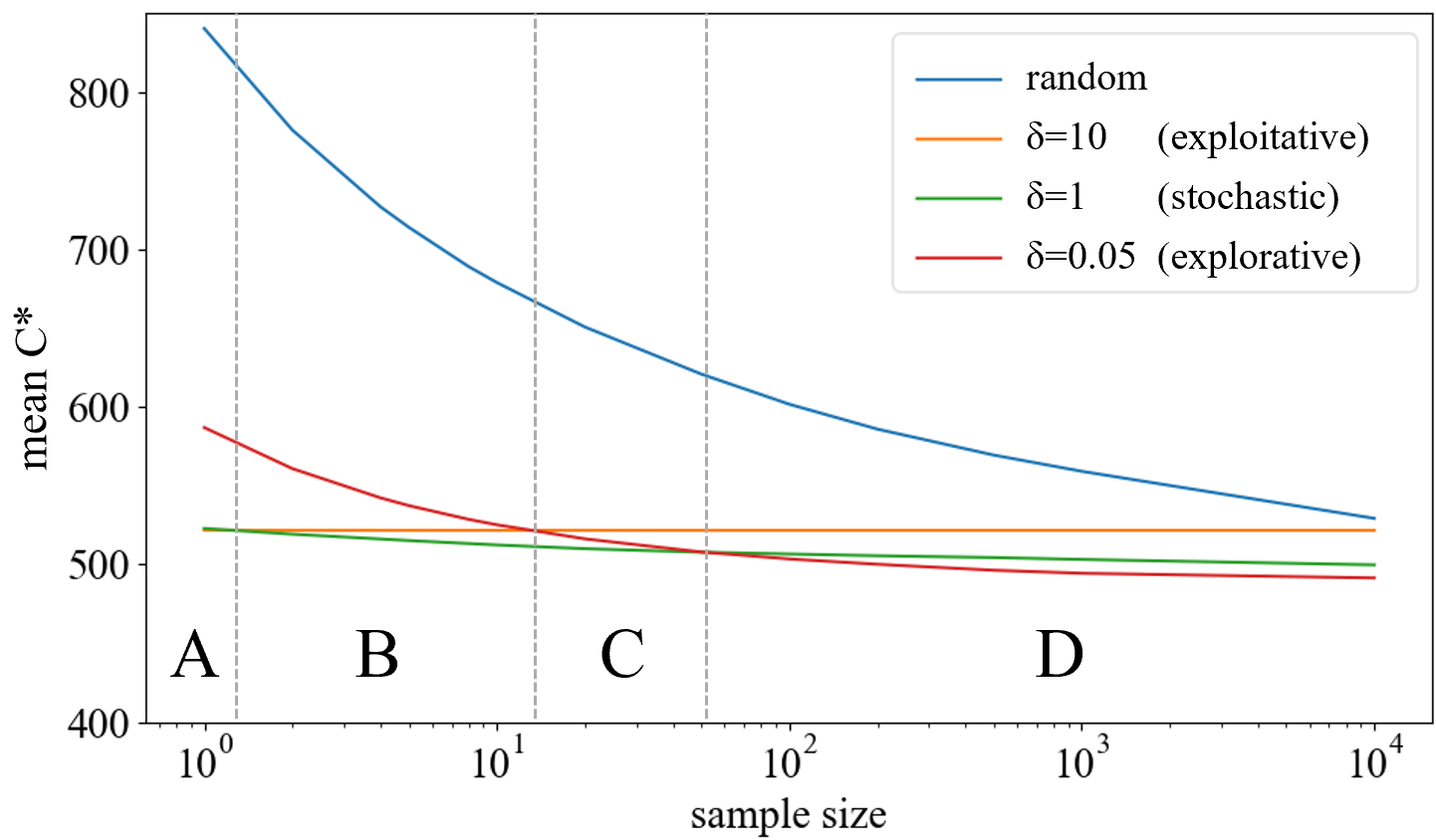}
  \caption{Minimal makespans over sampling size for different sampling strategies}
  \label{fig:hypothesis_validation}
\end{figure}

To empirically test our hypotheses illustrated in Figure~\ref{fig:hypothesis}, we create 10,000 solutions ($C^{*}_{10k}$) on ten 6x6 JSSP instances with our \textit{L2G} base model and choose four sampling parameterizations: 
\begin{enumerate}
    \item random sampling, which draws actions from a uniform distribution; 
    \item $\delta = 1$ as baseline of how sampling is usually performed; 
    \item $\delta = 0.05$ as extreme example of explorative sampling; 
    \item $\delta = 10$ as extreme example of exploitative sampling. 
\end{enumerate}  
Figure \ref{fig:hypothesis_validation} displays the averaged minimal makespan achieved over different sample sizes (logarithmic scale). To make up for statistical outliers in small sample sizes, we average the minimal makespans achieved over multiple sampling runs. Using Python array slicing notation, this can be formulated as in Equation \ref{eq:slidingwindow},
\begin{eqnarray}\label{eq:slidingwindow}
\frac{1}{10,000/s}\sum_{i=0}^{10,000/s}min\left( C^{*}_{10k}[i:s(i+1)] \right) ,
\end{eqnarray}
where s is the sample size.

In Figure \ref{fig:hypothesis_validation}, we divide the plot into four zones at interesting line intersections. In zone A, which in this example only includes a sample size of one, strongly exploitative sampling yields the best solutions. In zones B and C, stochastic sampling with $\delta = 1$ yields better makespans than strongly exploitative or explorative sampling. For sampling sizes $< 50$, strongly explorative sampling with $\delta = 0.05$ performs better than stochastic sampling. The switch from zone B to C indicates, where explorative sampling outperforms exploitative sampling. The results in Figure \ref{fig:hypothesis_validation} thereby verify our hypothesis.

\begin{table}
\centering
\begin{tabular}{lll p{0.7cm} p{0.7cm} p{0.7cm}}
base      &problem size & \multicolumn{3}{c}{sample size} \\
    \hline    \hline
    &      & 32    & 128   & 512     \Tstrut    \\
    \cline{2-5}
    \multirow{4}{*}{L2G} & 6x6    & 0.30 & 0.10 & 0.11 \Tstrut \\
                        & 15x15  & 0.70 & 0.86 & 0.43 \\
    & 20x20  & 1.25 & 1.46 & 0.80 \\
    & 100x20 & 0.58 & 0.43 & n.a. \Bst \\
    
    \cline{2-5}
   \multirow{4}{*}{L2D} & 6x6    & 2.06 & 1.37 & 0.93 \Tstrut \\
   & 15x15  & 12.25 & 6.02 & 3.32 \\
    & 20x20  & 10.35 & 10.38 & 10.05 \\
    & 100x20 & 14.0 & 10.88 & n.a. \\
\end{tabular}
\caption{Best found delta values for different base models, problem sizes and sample sizes}
\label{delta_values}
\end{table}

\begin{figure*}[h]
  \centering
  \includegraphics[width=\linewidth]{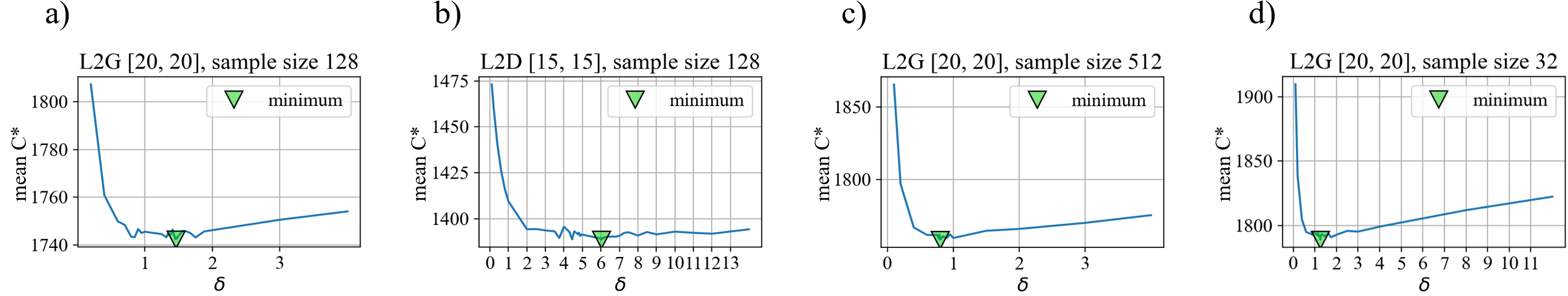}
  \caption{Examples of results in the $\delta$ value search algorithm with found minima}
  \label{fig:delta_search_examples}
\end{figure*}

\subsection{Optimal $\delta$-Values}
 In the following we present the results of finding and using the optimal parameterizations of our action sampling approach, adhering to the sample size of 128 used in \cite{Iklassov.09.06.2022b} for comparison and adding 32 and 512 as significantly smaller and larger examplary sample sizes. We omitted experiments on 100x20 JSSP with 512 samples, since we expect little additional insights compared to the considerable required computational times for solving large instances. Table \ref{delta_values} shows the results obtained from our parameter search algorithm per base model, problem size, and sample size. 

The results show two general trends. Firstly, in accordance with the above hypothesis, larger $\delta$-values are favorable for smaller sample sizes. Secondly, more exploitative sampling seems favorable with increasing problem size. However, we also observe exceptions to these general trends. For \textit{L2G} and problem sizes 15x15 and 20x20, there is a small increase of found $\delta$-values from sample size 32 to 128, before they decrease again for 512. Also for \textit{L2G}, the trend of increasing $\delta$-values with increasing problem sizes discontinues for problem size 100x20. Overall, the found $\delta$-values of \textit{L2G} are smaller in comparison to \textit{L2D}. These discontinuing trends, exceptions, and differences between base models lead us to believe that analytical approaches to finding $\delta$-values are intractable and they underscore the necessity to determine the sampling parameterization empirically and individually for any trained model and sample size.

For a deeper analysis, we examine the function $C^{*}(\delta)$ which was tested on the evaluation instances during $\delta$-value search. Noteworthy representative examples of this function are presented in Figure \ref{fig:delta_search_examples}. Across all examples, the functions satisfy the assumed smoothness and the hypothesis of a single minimum is valid, although some noise exists. Our iterative algorithm to find these minima converges towards reasonable minima, which are marked by the green triangles. Note that the small local minima, e.g. in Figure \ref{fig:delta_search_examples}a) at $\delta \approx 0.8$, could render gradient-based methods ineffective. However, as Figure \ref{fig:delta_search_examples}a) also exemplifies, $\delta$ can have a minimal overall effect, and in cases such as Figure \ref{fig:delta_search_examples}b), the function plateaus across large $\delta$ ranges. In light of these observations it is likely that the rare exceptions to decreasing $\delta$-values with increasing sample size in \textit{L2G} in Table \ref{delta_values}, which were discussed above, are relics of statistical noise.

\newcommand{\mybar}[1]{%
    \ifdim#1pt>0pt
        \settowidth{\mybarlen}{th}
        \textcolor{lightgreen}{\rule[-0.5ex]{#1\mybarlen}{2.0ex}}%
    \else
        \settowidth{\mybarlen}{th}%
        \hspace*{#1\mybarlen}%
        \textcolor{lightred}{\rule[-0.5ex]{-#1\mybarlen}{2.0ex}}%
    \fi
}

\newcolumntype{L}[1]{>{\raggedright\arraybackslash}p{#1\linewidth}|}

\begin{table*}[tp]
\centering
\begin{tabular}{clllllll} 
sample  size        & base                 & problem size & ours   & stochastic & deterministic & \multicolumn{2}{l}{improvement}  \\ 
\hline\hline
\multirow{8}{*}{32}  & \multirow{4}{*}{L2G} & 6x6   \Tstrut       & 496.0  & 505.5      & 517.0         & 1.9\%  &   \mybar{1.9}                      \\
                     &                      & 15x15        & 1310.4 & 1311.4     & 1364.5        & 0.1\%  &   \mybar{0.1}                      \\
                     &                      & 20x20        & 1782.9 & 1782.3     & 1866.2        & 0.0\%  &   \mybar{0.0}                      \\
                     &                      & 100x20  \Bst     & 5759.6 & 5764.1     & 5867.3        & 0.1\%  &   \mybar{0.1}                      \\ 
\cline{3-8}
                     & \multirow{4}{*}{L2D} & 6x6 \Tstrut         & 519.3  & 516.1      & 577.1         & -0.6\% &   \mybar{-0.6}                      \\
                     &                      & 15x15        & 1406.9 & 1434.6     & 1534.7        & 1.9\%  &    \mybar{1.9}                     \\
                     &                      & 20x20        & 1891.4 & 1957.4     & 2012.5        & 3.4\%  &   \mybar{3.4}                      \\
                     &                      & 100x20  \Bst      & 5865.8 & 5928.4     & 5996.7        & 1.1\%  &   \mybar{1.1}                      \\ 
\cline{2-8}
\multirow{8}{*}{128} & \multirow{4}{*}{L2G} & 6x6  \Tstrut        & 491.7  & 503.6      & 517.0         & 2.4\%  &   \mybar{2.4}                      \\
                     &                      & 15x15        & 1300.4 & 1300.9     & 1364.5        & 0.0\%  &   \mybar{0.0}                      \\
                     &                      & 20x20        & 1769.3 & 1767.6     & 1866.2        & -0.1\% &   \mybar{-0.1}                      \\
                     &                      & 100x20       & 5733.6 & 5745.1     & 5867.3        & 0.2\%  &   \mybar{0.2}                      \\ 
\cline{3-8}
                     & \multirow{4}{*}{L2D} & 6x6  \Tstrut        & 504.8  & 506.2      & 577.1         & 0.3\%  &   \mybar{0.3}                      \\
                     &                      & 15x15        & 1385.7 & 1410.1     & 1534.7        & 1.7\%  &   \mybar{1.7}                      \\
                     &                      & 20x20        & 1865.8 & 1927.6     & 2012.5        & 3.2\%  &   \mybar{3.2}                      \\
                     &                      & 100x20 \Bst      & 5869.8 & 5899.7     & 5996.7        & 0.5\%  &   \mybar{0.5}                      \\ 
\cline{2-8}
\multirow{6}{*}{512} & \multirow{3}{*}{L2G} & 6x6  \Tstrut        & 489.9  & 501.6      & 517.0         & 2.3\%  &   \mybar{2.3}                      \\
                     &                      & 15x15        & 1281.6 & 1291.7     & 1364.5        & 0.8\%  &   \mybar{0.8}                      \\
                     &                      & 20x20 \Bst       & 1752.0 & 1750.6     & 1866.2        & -0.1\% &   \mybar{-0.1}                      \\ 
\cline{3-8}
                     & \multirow{3}{*}{L2D} & 6x6 \Tstrut         & 498.0  & 498.7      & 577.1         & 0.1\%  &   \mybar{0.1}                      \\
                     &                      & 15x15        & 1369.1 & 1386.9     & 1534.7        & 1.3\%  &   \mybar{1.3}                      \\
                     &                      & 20x20 \Bst       & 1842.2 & 1879.6     & 2012.5        & 2.0\%  &   \mybar{2.0}                     
\end{tabular}
\caption{Average $C^{*}$ values on 100 generated test instances per problem size}
\label{own_instance_table}
\end{table*}

\newcommand{\mybartwo}[1]{%
    \ifdim#1pt>0pt
        \settowidth{\mybarlen}{th}
        \textcolor{lightgreen}{\rule[-0.5ex]{#1\mybarlen}{2.0ex}}%
    \else
        \settowidth{\mybarlen}{thl}%
        \hspace*{#1\mybarlen}%
        \textcolor{lightred}{\rule[-0.5ex]{-#1\mybarlen}{2.0ex}}%
    \fi
}

\begin{table*}[bp]
\centering
\begin{tabular}{lllllllll}
s.-size        & base                 & p.-size & ours & stochastic & deterministic & \multicolumn{2}{c}{improvement} & optimal  \\ 
\hline\hline
                     &                      &              & C* (opt. gap)      & C* (opt. gap)                   & C* (opt. gap)                   & \multicolumn{2}{c}{ours vs. stoch.}              &    C      \\ 
\cline{3-9}
\multirow{6}{*}{32}  & \multirow{3}{*}{L2G} & 15x15        & 1345.9 (9.5\%)          & 1345.1 (9.5\%)                 & 1417.4 (15.3\%)                   & -0.1\%     & \mybartwo{-0.1}                  & 1228.9  \Tstrut \\
                     &                      & 20x20        & 1858.9 (14.9\%)         & 1858.7 (14.9\%)                & 1935.3 (19.7\%)                   & 0.0\%      & \mybartwo{0.0}                   & 1617.3   \\
                     &                      & 100x20       & 5759.7 (7.3\%)          & 5784.6 (7.8\%)                 & 5878.4 (9.6\%)                    & 0.4\%      & \mybartwo{0.5}                   & 5365.7 \Bst  \\ 
\cline{2-9}
                     & \multirow{3}{*}{L2D} & 15x15        & 1452.5 (18.2\%)         & 1465.9 (19.3\%)                & 1530.5 (24.5\%)                   & 0.9\%      & \mybartwo{1.1}                   & 1228.9 \Tstrut  \\
                     &                      & 20x20        & 1953.2 (20.8\%)         & 2027.1 (25.3\%)                & 2081.8 (28.7\%)                   & 3.6\%      & \mybartwo{4.6}                   & 1617.3   \\
                     &                      & 100x20       & 5928.6 (10.5\%)         & 5983.1 (11.5\%)                & 6089 (13.5\%)                   & 0.9\%      & \mybartwo{1.0}                   & 5365.7  \Bst \\ 
\cline{2-9}
\multirow{6}{*}{128} & \multirow{3}{*}{L2G} & 15x15        & 1340.3  (9.1\%)          & 1340.9 (9.1\%)                 & 1417.4 (15.3\%)                   & 0.0\%      & \mybartwo{0.0}                   & 1228.9   \Tstrut\\
                     &                      & 20x20        & 1846.4  (14.2\%)         & 1837.4 (13.6\%)                & 1935.3 (19.7\%)                   & -0.5\%     & \mybartwo{-0.6}                  & 1617.3   \\
                     &                      & 100x20       & 5751.1  (7.2\%)          & 5766.5 (7.5\%)                 & 5878.4 (9.6\%)                    & 0.3\%      & \mybartwo{0.3}                   & 5365.7   \Bst \\ 
\cline{3-9}
                     & \multirow{3}{*}{L2D} & 15x15        & 1416.2  (15.2\%)         & 1442   (17.3\%)                & 1530.5 (24.5\%)                   & 1.8\%      & \mybartwo{2.1}                   & 1228.9  \Tstrut  \\
                     &                      & 20x20        & 1943.4  (20.2\%)         & 2002.7 (23.8\%)                & 2081.8 (28.7\%)                   & 3.0\%      & \mybartwo{3.7}                   & 1617.3   \\
                     &                      & 100x20       & 5888.2  (9.7\%)          & 5930.4 (10.5\%)                & 6089   (13.5\%)                   & 0.7\%      & \mybartwo{0.8}                   & 5365.7  \Bst \\ 
\cline{2-9}
\multirow{4}{*}{512} & \multirow{2}{*}{L2G} & 15x15        & 1323.1  (7.7\%)          & 1333.7 (8.5\%)                 & 1417.4 (15.3\%)                   & 0.8\%      & \mybartwo{0.9}                   & 1228.9  \Tstrut\\
                     &                      & 20x20        & 1825.5  (12.9\%)         & 1828.9 (13.1\%)                & 1935.3 (19.7\%)                   & 0.2\%      & \mybartwo{0.2}                   & 1617.3  \Bst \\ 
\cline{3-9}
                     & \multirow{2}{*}{L2D} & 15x15        & 1409.0    (14.7\%)         & 1427.6 (16.2\%)                & 1530.5 (24.5\%)                   & 1.3\%      & \mybartwo{1.5}                   & 1228.9  \Tstrut \\
                     &                      & 20x20        & 1895.3  (17.2\%)         & 1946.2 (20.3\%)                & 2081.8 (28.7\%)                   & 2.6\%      & \mybartwo{3.1}                   & 1617.3  
\end{tabular}
\caption{Results on Taillard Instances}
\label{ta_instance_results}
\end{table*}

\subsection{Performance Improvements}

\textbf{Test Instances.} The effectiveness of the obtained parameterization is tested on 100 test instances. The average $C^{*}$ values are presented in Table \ref{own_instance_table}. We report the results for $\delta$-sampling (\textit{ours}), using stochastic sampling as in \citet{Iklassov.09.06.2022b} (\textit{stochastic}) as a baseline, both for three sample sizes, and the deterministic solution (\textit{deterministic}). Relative improvements between ours and stochastic sampling ($(stochastic - ours)/stochastic$) are depicted in the last column. 

Both sampling methods find significantly better solutions than the deterministic solution, confirming previous results \cite{Iklassov.09.06.2022b,WouterKool.2018, Orseau.}. Our parameterized method outperforms stochastic sampling in 17 out of 22 cases with up to 3.2\% improvement. Notably, improvements are achieved across all tested sample sizes. The most notable exception is $\delta$ sampling for the \textit{L2D} base model on the 6x6 JSSP, which we partially attribute to statistical uncertainty given the small sample size of 32. An exception that persists across sample sizes is \textit{L2G} on the 20x20 JSSP, where $\delta$-sampling performs slightly worse than the stochastic baseline. 
Overall, we observe larger performance differences where $\delta$ deviates more from one, i.e. stochastic sampling. This implies that our parameter search is effective in finding stronger manipulations where they are useful. On the other hand, constellations where $\delta$-values are close to one and reside within plateaus of the function $C^{*}(\delta)$, as shown in Figure \ref{fig:delta_search_examples} c) and d), only minor performance differences can be obtained, as observed for \textit{L2G} on the 20x20 JSSP.

\begin{figure*}[ht!]
  \centering
  \includegraphics[width=0.8\linewidth]{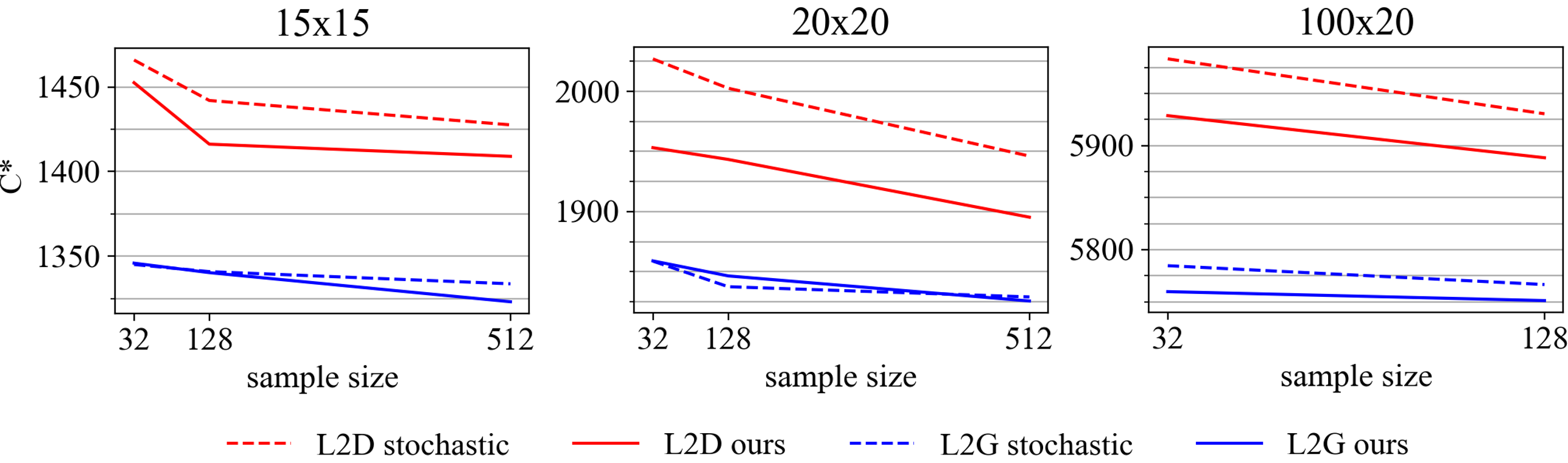}
  \caption{Comparison of sampling methods over sample sizes}
  \label{fig:own_vs_stoch}
\end{figure*}

To understand those cases where very similar average $C^{*}$ are generated, we analyze the resulting schedule differences. Within the schedules of \textit{L2G} on the 15x15 JSSP with 128 samples, the most striking example, we counted in how many cases the same schedules, measured by the same $C^{*}$, were found per instance. Surprisingly, $C^{*}$ was equal in only 4/100 test instances. This shows that even though the average values of $C^{*}$ were less than $0.00\%$ apart, different solutions were found, such that $\delta$ still had a significant impact. In contrast, for 6x6 JSSPs of \textit{L2G} with a sample size of 128, we obtained the same $C^{*}$ in roughly one third of the instances. This is most likely an effect of the smaller total solution space compared to the other problem sizes. The conclusion is that even if only a few different solutions are found, they can still have a significant impact, as the $1.9\%$ improvement demonstrates. Therefore, our sampling method is effective for small problem sizes with smaller solution spaces, too.

\textbf{Benchmark Instances.} We further report the results on benchmark instances of \citet{Taillard.1993}, which we abbreviate as TA instances, in Table \ref{ta_instance_results}. In comparison, there are only ten instances per size, but the optimal solutions are known, enabling us to additionally provide the relative gap (\textit{opt. gap}) for reference. The effects of sampling methods are fundamentally similar. As before, both sampling approaches outperform the deterministic solution. While the results on \textit{L2D} indicate even greater improvements through our sampling method, the results on \textit{L2G} are more ambiguous. We attribute the differences of the above results to the fact that only ten TA instances exist per problem size, introducing a larger statistical error. Note that overall we find slightly larger makespans than those reported in \cite{Iklassov.09.06.2022b}, as we could not reproduce the reported performance on benchmarks with our trained model. However, our contribution is a relative improvement per model and sample size applicable to any learned construction heuristic, not the creation of a new learning strategy or network architecture. For details on the most recent absolute results and a comprehensive comparison of learning-based, priority-rule-based or heuristic methods, we refer the interested reader to \cite{Iklassov.09.06.2022b}, \cite{corsini2024selflabeling} and \cite{.2022b}.

In Figure \ref{fig:own_vs_stoch}, relative improvements on TA instances through our adjusted sampling method can be compared to improvements achievable by increasing sample sizes. In many cases, we find that the achieved $C^{*}$ through $\delta$-sampling is similar or better than the $C^{*}$ achieved via stochastic sampling with a four times larger sample size. In practical terms, the same results can be achieved with a quarter of the computational budget through better sampling in these cases, rendering $\delta$-sampling four times more efficient.

\section{Conclusion and Outlook}
In this paper, we proposed a sampling method for the more effective usage of trained DRL-based construction heuristics for the JSSP. We showed that an optimal balance between exploration and exploitation during inference exists, which is specific to the trained model and the chosen sample size. Based on this insight, a method was developed that allows the tradeoff between exploration and exploitation to be easily and smoothly parameterized during solution generation of learned construction heuristics. The effectiveness of this method was experimentally evaluated and showed promising results across different base models and computational budgets on common JSSP sizes. We believe that our approach is neither limited to the studied base models nor the application to the JSSP, but can be used to improve any other learned construction heuristic on both the JSSP and other related combinatorial optimization problems.

In the future, we would like to transfer the insights gained into training protocols that steer agents to find the best possible solutions given a previously set computational budget to enforce a learned exploration-exploitation tradeoff. In addition, we will combine our sampling method with Monte Carlo Tree Search (MCTS) for a more effective policy-guided search during inference.

\bibliography{main}

\end{document}